\documentclass[conference]{IEEEtran}
\IEEEoverridecommandlockouts

\usepackage{cite}
\usepackage{graphicx} 
\usepackage{soul}
\usepackage{url}
\usepackage{adjustbox} 
\usepackage{booktabs}
\usepackage{graphicx}
\usepackage{algorithm}
\usepackage{algorithmic}
\usepackage[table]{xcolor}
\usepackage{adjustbox}
\usepackage{lscape}
\usepackage{colortbl}
\usepackage{subfig}
\usepackage{subcaption}
\usepackage{multirow}
\usepackage{pifont}
\usepackage{subcaption}
\usepackage{float}
\usepackage{amsmath,amssymb,amsfonts}
\usepackage{algorithmic}
\usepackage{graphicx}
\usepackage{textcomp}
\usepackage{xcolor}
\def\BibTeX{{\rm B\kern-.05em{\sc i\kern-.025em b}\kern-.08em
    T\kern-.1667em\lower.7ex\hbox{E}\kern-.125emX}}

\begin{document}

\title{CausalGeD: Blending Causality and Diffusion for Spatial Gene Expression Generation}
\author{
    \IEEEauthorblockN{
        \textsuperscript{1}Rabeya Tus Sadia, 
        \textsuperscript{2}Md Atik Ahamed, 
        \textsuperscript{3}Qiang Cheng
    } 
    \IEEEauthorblockA{
        \textsuperscript{1,2,3}\textit{Department of Computer Science, \textsuperscript{3}Institute for Biomedical Informatics} \\ 
        \textit{University of Kentucky, Lexington, Kentucky, USA} \\ 
        \{rabeya.sadia, atikahamed, qiang.cheng\}@uky.edu
    }
    \thanks{\textsuperscript{3}Corresponding author: Qiang Cheng}
}

\maketitle  

\begin{abstract}
The integration of single-cell RNA sequencing (scRNA-seq) and spatial transcriptomics (ST) data is crucial for understanding gene expression in spatial context. Existing methods for such integration have limited performance, with structural similarity often below 60\%, We attribute this limitation to the failure to consider causal relationships between genes. We present CausalGeD, which combines diffusion and autoregressive processes to leverage these relationships. By generalizing the Causal Attention Transformer from image generation to gene expression data, our model captures regulatory mechanisms without predefined relationships. Across 10 tissue datasets, CausalGeD outperformed state-of-the-art baselines by 5- 32\% in key metrics, including Pearson's correlation and structural similarity, advancing both technical and biological insights.
\end{abstract}

\maketitle

\section{Introduction}
Integrating spatial transcriptomics (ST) with single-cell RNA sequencing (scRNA-seq) combines spatial context with high-resolution gene expression, enhancing cell-type mapping, tissue microenvironment analysis, and noise reduction. This integration improves spatial data deconvolution, aids missing data recovery, and enables applications in tumor heterogeneity, disease progression, and tissue organization.

Causal relationships between genes are critical because they capture the underlying biological mechanisms of gene regulation. When one gene affects the expression of another through direct or indirect regulatory pathways, traditional correlation-based methods may miss these directional dependencies. For example, in transcription factor networks, the expression of upstream regulators directly influences their downstream targets, creating a temporal and causal chain of events. By incorporating these causal relationships, our model can better predict gene expression patterns by learning not just what genes are expressed together, but how they influence each other's expression levels.

\begin{figure}[htp]
    \centering
    \includegraphics[width=0.75\linewidth]{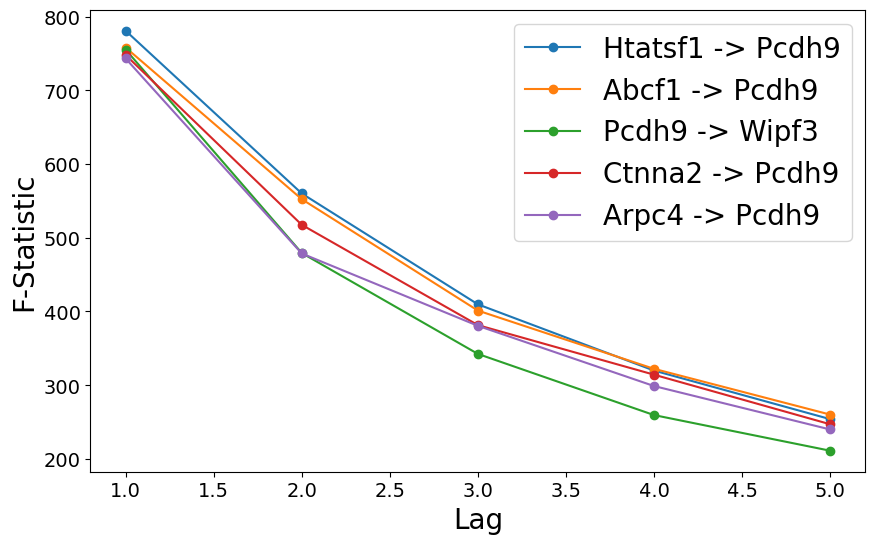}
    \caption{Top 5 gene pairs with strongest Granger causality relationships from randomly selected MC data (Section 4). All pairs show highly significant causal relationships (p-values < 1e-16), highlighting the importance of modeling gene-gene causality.}
    \label{fig:causality}
\end{figure}

Machine learning approaches have been developed for scRNA-seq and ST integration, including SpaGE, SpatialScope, novoSpaRc, and stDiff~\cite{SpaGE,SpatialScope,novoSpaRc,stDiff}. Our Granger causality analysis (Figure~\ref{fig:causality}) reveals that these methods overlook important causal relationships between genes, leading to consistently limited performance (correlation coefficients typically below 60\%)~\cite{SpatialScope, li2024spadit}.

Incorporating causal relationships in this context is challenging, because standard causality measures have not been widely applied to spatial omics. For example, Granger causality, a statistical framework for identifying temporal dependencies and directional relationships in gene expression data, provides valuable insights into gene regulatory mechanisms~\cite{yao2015prior}. While traditionally used for temporal data, Granger causality effectively captures regulatory relationships in gene expression through analysis of expression level dependencies, as validated by our statistical tests showing significant F-statistics (\( p < 0.001 \)) across multiple gene pairs. While this analysis highlights the superiority of causality-driven methods over traditional correlation-based approaches in understanding gene expression dynamics, incorporating these insights into gene expression prediction remains challenging.

Traditional causal analysis typically requires predefined regulatory relationships, which is overly restrictive when exploring general ST and scRNA-seq data integration. This raises the question: How can we effectively incorporate causal relationships into the integration of ST and scRNA-seq data without requiring extensive prior knowledge while maintaining computational feasibility?

To address these challenges, we develop a novel causality-aware model that incorporates causal relationships while predicting gene expression values, aiming to improve both biological interpretability and predictive accuracy. Building on the success of diffusion models in natural language processing and computer vision~\cite{rombach2022high, li2022diffusion, DiT}, we combine causality modeling with the diffusion process to create an architecture that effectively facilitates scRNA-seq and ST integration.

Our contributions include:\begin{itemize}
\item We propose a novel framework, CausalGeD, which combines diffusion and autoregressive models to leverage causal relationships between genes for integrating scRNA-seq and ST data, particularly generating 
missing values at ST spots.
\item We develop a Causality Aware Transformer (CAT) module to blend autoregression with diffusion for capturing long-range causal relationships in gene expression data, enabling effective handling of high-dimensional continuous data without spatial dependencies.
\item Through extensive experiments across 10 diverse tissue datasets, our model demonstrates superior performance compared to state-of-the-art baselines, validating its practical utility.
\end{itemize}

These advances not only improve integration accuracy but also provide deeper insights into gene regulatory mechanisms in spatial context.
\begin{figure*}[htb] 
    \centering{
    \includegraphics[width=0.9\linewidth]{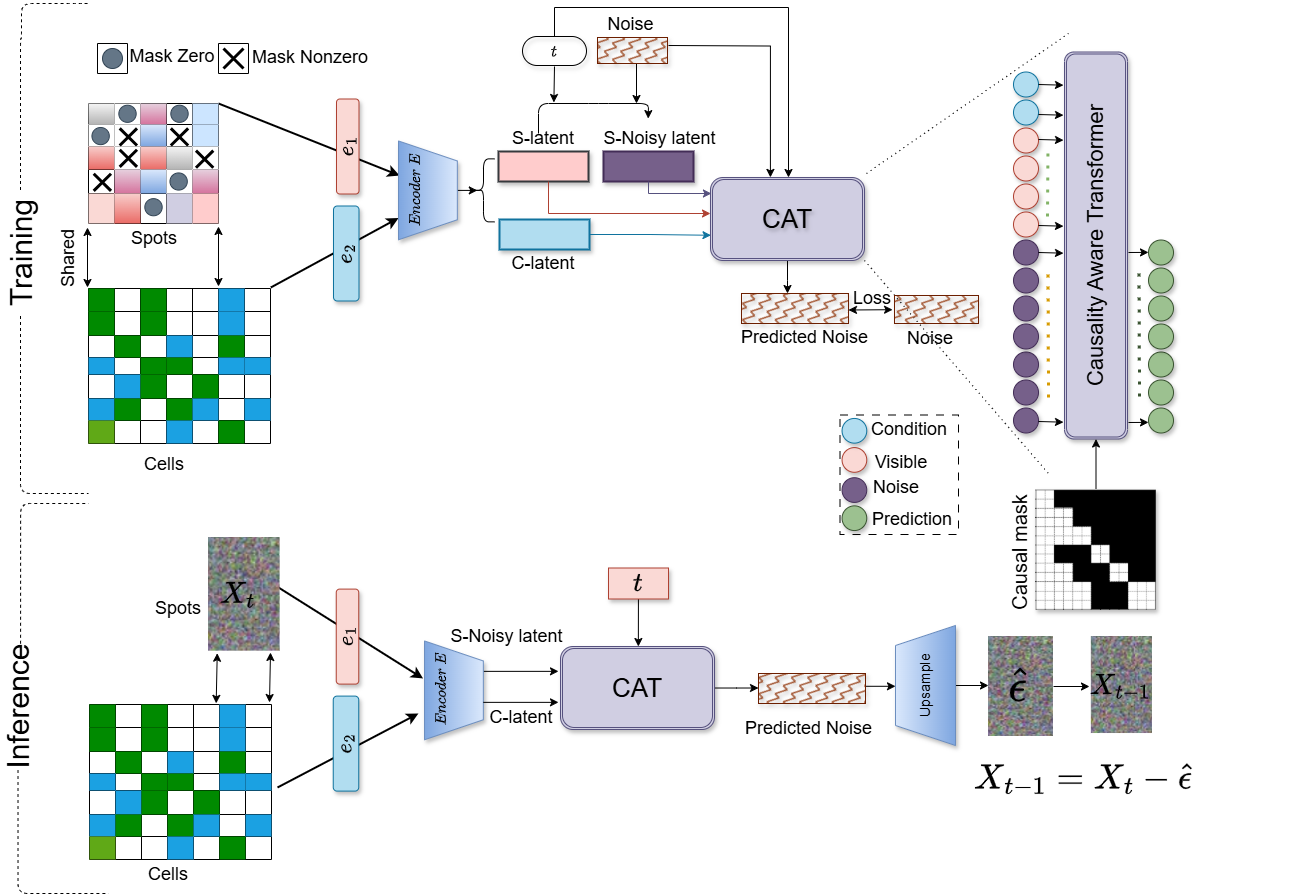}
    \caption{Architecture overview of CausalGeD. The framework comprises both training and inference processes, implemented through key components: latent space module, noisy input construction, diffusion-autoregression integration, causal attention mask used in causality-aware Transformer (CAT), and an inference module for processing noisy inputs.}
    \label{fig:method}}
\end{figure*}
\section{Related Works}
Recent approaches for integrating ST and scRNA-seq data can be broadly categorized into several groups based on their methodological foundations:

\textbf{Alignment and Mapping-based Methods:} Tangram~\cite{Tangram} employs deep learning to align scRNA-seq data with spatial data by learning a mapping between the two modalities. This approach enables high-resolution spatial mapping of cell types and states. Similarly, SpaGE~\cite{SpaGE} enhances spatial gene expression by learning gene-gene relationships from scRNA-seq data and transferring this information to spatial data.

\textbf{Deep Generative Models:} Several methods leverage advanced generative modeling approaches. scVI~\cite{scVI} provides a deep generative framework for single-cell transcriptomics that can be extended to spatial data integration. Building upon this foundation, SpatialScope~\cite{SpatialScope} employs deep generative models specifically designed to integrate spatial and single-cell transcriptomics data.

\textbf{Transport-based Methods:} Methods such as SpaOTsc~\cite{SpaOTsc} and novoSpaRc~\cite{novoSpaRc} utilize optimal transport theory. SpaOTsc infers spatial and signaling relationships between cells from transcriptomic data, while novoSpaRc offers flexible spatial reconstruction of gene expression using optimal transport principles.

\textbf{Reference-based Enhancement:} stPlus~\cite{stPlus} takes a reference-based approach to enhance ST data accuracy. This method focuses on improving the resolution and quality of spatial data by leveraging information from scRNA-seq references.

\textbf{Diffusion-based Models:} Recent approaches have explored diffusion models for this integration task. stDiff~\cite{stDiff} employs a diffusion model for imputing ST through single-cell data. SpaDiT~\cite{li2024spadit} builds upon this framework by incorporating transformer architectures for spatial gene expression prediction, demonstrating the potential of combining diffusion models with attention mechanisms.

In summary, while existing methods for integrating ST and scRNA-seq data have made significant progress, they generally lack explicit modeling of causal relationships between genes, a critical factor in biological systems. Alignment and mapping-based methods like Tangram and SpaGE excel in spatial alignment but rely heavily on predefined relationships. Deep generative models, including scVI and SpatialScope, provide flexible frameworks but often overlook biological dependencies in the data. Transport-based approaches such as SpaOTsc and novoSpaRc emphasize spatial reconstruction but fail to capture gene regulatory dynamics. Even recent diffusion-based models like stDiff and SpaDiT, which integrate advanced generative techniques, focus primarily on spatial continuity rather than causal dependencies. In contrast, our approach, CausalGeD, uniquely incorporates gene-gene causal relationships through a novel combination of diffusion processes and autoregressive modeling. This enables more accurate and biologically meaningful integration, addressing the limitations of prior methods and advancing the field both technically and scientifically.

\section{Methodology}
\subsection{Overview of Our Model}
Our goal is to leverage scRNA-seq data to predict unmeasured gene expression in ST data by integrating them into latent spaces. Given gene expression matrices $X_{st} = \{x_{st}^i\}_{i=1}^n \in \mathbb{R}^{n \times p}$ from ST and $X_{sc} = \{x_{sc}^j\}_{j=1}^m \in \mathbb{R}^{m \times q}$ from scRNA-seq (where $n$, $m$ are gene counts and $p$, $q$ are spot/cell counts), we aim to predict gene locations and expression values missing in ST data.

Our key insight is incorporating the commonly existing but overlooked causal relationships between genes. As demonstrated by our Granger causality analysis (Figure \ref{fig:causality}), these relationships are statistically significant across datasets, with high F-statistics and near-zero p-values validating their strength.

We develop CausalGED to utilize causal relationships within a diffusion architecture (Figure \ref{fig:method}), operating in two stages: training, which employs an encoder and a causality-aware transfer (CAT) module for forward diffusion, and inference, which integrates these components with a decoder for reverse diffusion to generate expression values. The encoder transforms input data into C-latent and S-latent representations, enabling the CAT module to model causal dependencies. The diffusion process refines gene expression predictions, while an autoregressive component preserves regulatory relationships, ensuring both local and global expression patterns are maintained.

We summarize our notations as follows: $x_{st}^i \in \mathbb{R}^p$: spatial transcriptomics data for gene $i$;  $x_{sc}^j \in \mathbb{R}^q$: scRNA-seq data for gene $j$; $E$: a two-headed encoder $E: \mathbb{R}^p$ or $\mathbb{R}^q \rightarrow \mathbb{R}^d$, where $d$ is the latent dimension; $\hat{x}_{st}^i$: latent representation of spatial data where $\hat{x}_{st}^i = E(x_{st}^i)$;
$\hat{x}_{st}^{i,t}$: noisy version of latent representation at diffusion step $t$;
$\kappa_s$: subset of genes processed in autoregressive step $s$.

\subsection{Modules of Our Architecture}
We provide a detailed description of the modules in our model. Our proposed CausalGeD model contains the following modules.

\paragraph{\bf Latent space representation:}
To achieve latent space representation, we use a two-headed encoder $E$, as illustrated in Figure~\ref{fig:method}. One head, $e_1$, captures $x_{st}^i$, while the other head, $e_2$, captures $x_{sc}^j$, depending on the input dimensions $p$ and $q$. The encoder transforms $x_{st}^i \in \mathbb{R}^p$ into a latent space representation $\hat{x}_{st}^i \in \mathbb{R}^d$, referred to as S-latent in Figure~\ref{fig:method}, and similarly transforms $x_{sc}^j \in \mathbb{R}^q$ into $\hat{x}_{sc}^j \in \mathbb{R}^d$. After obtaining these latent representations, we construct a noisy version of $\hat{x}_{st}^i$.

ST and scRNA-seq data differ fundamentally—ST provides spatial context at lower resolution, while scRNA-seq offers high-resolution expression without spatial information. A two-headed encoder ensures each modality is encoded separately, preserving unique features, reducing bias, and aligning representations in latent space for effective integration and cross-modal prediction.

\paragraph{\bf Noisy input construction:}
We construct the noisy input, referred to as S-Noisy latent in Figure~\ref{fig:method}, using a forward diffusion process. In this process, random noise is gradually added to $\hat{x}_{st}^i$ to form a Markov chain. With $t$ denoting a diffusion step that is also a noise level, each noisy version, $\hat{x}_{st}^{i,t}$, depends on its previous state, $\hat{x}_{st}^{i,t-1}$, modeled by a transition probability distribution:
\begin{equation}
\label{eq-transition}
    q(\hat{x}_{st}^{i,t} | \hat{x}_{st}^{i,t-1}) = \mathcal{N}(\hat{x}_{st}^{i,t}; \sqrt{1 - \beta_t} \, \hat{x}_{st}^{i,t-1}, \beta_t \mathbf{I}),
\end{equation}
where $\beta_t$ is the variance schedule parameter at step $t$, and $I$ is the identity matrix. The variance schedule $\{\beta_t\}_{t=1}^T$ is defined as a monotonically increasing sequence $\beta_1 < \beta_2 < \cdots < \beta_T$, controlling the gradual addition of noise. At each step $t$, the transition probability $q(\hat{x}_{st}^{i,t}|\hat{x}_{st}^{i,t-1})$ defines how the latent representation evolves. To obtain the complete forward process, we combine these step-wise transitions through the chain rule of probability, factorizing the joint distribution over all noise levels as:
\begin{equation} \label{eq:diffusion-factorization}
    q(\hat{x}_{st}^{i,0:T}) = q(\hat{x}_{st}^{i,0}) \prod_{t=1}^T q(\hat{x}_{st}^{i,t} | \hat{x}_{st}^{i,t-1}).
\end{equation}
{\bf{Diffusion Sampling Strategy}}: Our model employs a diffusion strategy that balances computational efficiency with performance. The forward diffusion process spans $T$ timesteps (default $T=2000$), where noise is introduced at each step $t$ according to variance schedule $\beta_t$. Instead of a fixed timestep for all autoregressive (AR) steps, we adopt a flexible sampling approach:\\
1. Base Strategy: For each AR step $s$, multiple diffusion timesteps are sampled to capture varying noise levels, enabling the model to learn gene dependencies at different scales. \\ 
2. Efficient Sampling: To optimize computation while preserving performance, we implement three ways:
Full sampling ($T$): uses all timesteps for maximum fidelity;  
Fractional sampling ($1/nT$): uses evenly spaced subsets ($n = 2,3,4,20$);  
Adaptive sampling: adjusts sampling density based on AR step importance.  

Empirical results (Table \ref{tab:timestep}) show that performance remains robust across configurations, with variations typically under 1\% PCC. This stability allows users to select sampling density based on computational constraints without compromising accuracy.

The sampling process integrates with AR steps via the transition probability distribution (Equation \ref{eq-transition}), where each AR step follows its own sampling schedule while reusing clean tokens from previous steps, ensuring computational efficiency and capturing gene regulatory dynamics at multiple scales.  

\paragraph{\bf Mixing Diffusion and Autoregression:}  
\begin{figure}[tbp]
    \centering
    \includegraphics[width=0.6\linewidth]{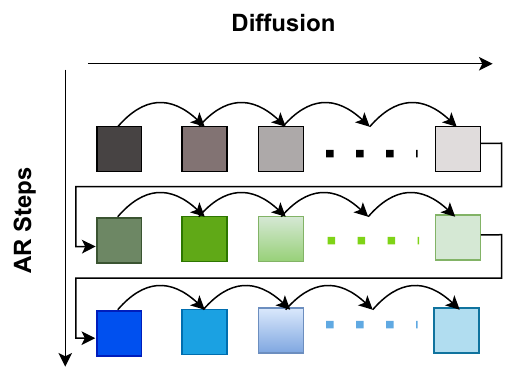} 
    \caption{Diffusion blended with autoregression. }
    \label{fig:cat_diff}
\end{figure}
While diffusion models excel at data generation, they typically do not capture dependencies that AR models naturally represent. We introduce the CAT module to combine both to mirror biological gene regulatory mechanisms. The CAT module processes genes in an autoregressive sequence that reflects how transcription factors and regulatory genes influence their downstream targets. This biological ordering is achieved through our causal attention mechanism, where earlier genes in the sequence can influence later ones but not vice versa, similar to how upstream regulators affect downstream targets in gene regulatory networks.

While our CAT architecture is inspired by~\cite{deng2024causal} for images, creating it for gene expression requires fundamental innovations. Unlike image generation that processes spatial dependencies through patches and class labels, gene expression requires handling high-dimensional continuous values and complex regulatory relationships. CAT addresses these challenges through three key components: (1) a modified conditional causality aware attention mechanism to capture continuous expression levels rather than discrete classes for images; (2) a token-based processing system treating compressed representations from each gene as a regulatory unit that is a token, unlike images where tokens represent sub-images; (3) a unified encoder integrates ST and scRNA-seq data with minimal additional parameters (two heads: $e_1$ and $e_2$), unlike multimodal image generation tasks (e.g., image captioning), which usually require separate encoders. This design enables CausalGeD to learn both strong direct regulatory relationships (e.g., transcription factor-target interactions) and weaker indirect ones (e.g., co-regulated genes).

The AR step sizes are randomly selected based on the AR step decay. For instance, in the first AR step, 2 tokens are chosen, followed by 2 tokens in the second step, and 3 tokens in the third step, as illustrated in Appendix Figure \ref{fig:attn_mask}. A more detailed explanation of this example is provided in the Appendix \ref{sec:causal_mask}. These step sizes were optimized by optimizing AR step decay to handle the non-uniform sparsity and noise characteristics of gene expression data while ensuring the model effectively learns dependencies across genes and spatial contexts. This integration enables the model to combine the strengths of diffusion and AR processes for biologically meaningful predictions. 

The core AR-diffusion integration leverages the following Eq.~\ref{eq:cd-factorization}, where $\kappa_s$ represents the subset of genes processed as a token during the $s$-th step of the AR process. Here, $1 \leq s \leq S$, with $S$ denoting the total number of AR steps.
\begin{equation}
\resizebox{\linewidth}{!}{$
q(\hat{x}_{st}^{i,0:T,\kappa_s} | \hat{x}_{st}^{i,0,\kappa_{1:s-1}}) = 
q(\hat{x}_{st}^{i,0,\kappa_s}) \prod_{t=1}^T 
q(\hat{x}_{st}^{i,t,\kappa_s} | \hat{x}_{st}^{i,t-1,\kappa_s},\hat{x}_{st}^{i,0,\kappa_{1:s-1}})
$}
\label{eq:cd-factorization}
\end{equation}
Here, $S$ denotes the total AR steps, and $\kappa_s$ identifies the subset of latent tokens processed at step $s$ (illustrated in Figure~\ref{fig:cat_diff}). During training, the model approximates $p_\theta(\hat{x}_{st}^{i,t-1,\kappa_s} | \hat{x}_{st}^{i,t,\kappa_s},\hat{x}_{st}^{i,0,\kappa_{1:s-1}})$ for all $t$ and $s$, using both noised tokens at the current AR step and clean tokens from previous steps.

\paragraph{\bf Causal Attention Mask:}
The causal attention mask ensures proper sequential dependencies by preventing $\hat{x}_{st}^{i,0,\kappa_{1:s-1}}$ from accessing $\hat{x}_{st}^{i,t,\kappa_s}$. This preserves causal dependencies across AR steps while ensuring each step only uses clean tokens from previous steps. During training, condition tokens are appended to clean tokens, with only the first $S-1$ AR steps utilizing clean tokens. The condition tokens remain unmasked throughout the process. As an illustration shown in Figure~\ref{fig:method}, it can be implemented with condition token length 2 and AR step sizes of 2, 2, and 3. Time-embedded noisy tokens are then appended and processed by the Transformer under mask guidance. For detailed mask formation, see Appendix~\ref{sec:causal_mask}. 

\paragraph{\bf Inference from Noisy Input:}
During inference, the model uses scRNA-seq data as a condition to guide the denoising process for generating ST data. The process begins by projecting both noisy input latent representations ($\hat{x}_{st}^{i,T}$) and conditional scRNA-seq data ($\hat{x}_{sc}^j$) into the latent space. The CAT module then predicts the noise component in the noisy latent input upsamples the prediction from latent to feature space ($\mathbb{R}^d \to \mathbb{R}^p$), followed by estimating the denoised latent representation ($\tilde{x}_{st}^{i,T-1}\in \mathbb{R}^p$). 

This reverse diffusion process iterates from $t = T$ to $1$, with CAT combining learned causal structure and conditional information from $\hat{x}_{sc}^j$ at each step. The final output represents the reconstructed ST data with predicted gene expression values.

\section{Experiment Analysis}
\label{sec_exp}

In this section, we first introduce the dataset and training details. We performed a comparative analysis of the CausalGeD method against the most widely recognized approaches to comprehensively evaluate its effectiveness in predicting gene expression. 

\begin{table*}[]
\centering
\caption{Comparison with baseline methods on the ten paired scRNA-seq and ST datasets format. The best results are represented in \textbf{Bold}. Four evaluation metrics (PCC, SSIM, RMSE, JS) are utilized to evaluate the methods.}
\label{baseline}
\resizebox{\textwidth}{!}{%
\begin{tabular}{l|c|c|c|c|c|c|c|c|c|c}
\toprule
PCC$\uparrow$ &
  MG &
  MH &
  MHPR &
  MVC &
  MHM &
  HBC &
  ME &
  MPMC &
  MC &
  ML \\ 
  \midrule
Tangram \cite{Tangram} &
  0.458±0.203 &
  0.523±0.116 &
  0.683±0.012 &
  0.623±0.117 &
  0.536±0.053 &
  0.703±0.142 &
  0.503±0.025 &
  0.727±0.026 &
  0.745±0.003 &
  0.714±0.056 \\
scVI \cite{scVI} &
  0.476±0.157 &
  0.446±0.157 &
  0.691±0.143 &
  0.594±0.023 &
  0.511±0.117 &
  0.656±0.005 &
  0.496±0.007 &
  0.716±0.014 &
  0.736±0.015 &
  0.637±0.001 \\
SpaGE \cite{SpaGE}&
  0.526±0.114 &
  0.438±0.163 &
  0.653±0.063 &
  0.603±0.107 &
  0.545±0.226 &
  0.639±0.025 &
  0.512±0.013 &
  0.753±0.066 &
  0.769±0.011 &
  0.653±0.007 \\
stPlus \cite{stPlus}&
  0.503±0.233 &
  0.401±0.037 &
  0.483±0.231 &
  0.574±0.059 &
  0.476±0.007 &
  0.597±0.111 &
  0.526±0.026 &
  0.689±0.007 &
  0.701±0.099 &
  0.699±0.014 \\
SpaOTsc \cite{SpaOTsc}&
  0.522±0.014 &
  0.485±0.107 &
  0.657±0.002 &
  0.629±0.147 &
  0.496±0.018 &
  0.587±0.107 &
  0.547±0.006 &
  0.734±0.201 &
  0.738±0.064 &
  0.723±0.005 \\
novoSpaRc \cite{novoSpaRc}&
  0.563±0.158 &
  0.567±0.252 &
  0.613±0.146 &
  0.656±0.037 &
  0.515±0.003 &
  0.647±0.122 &
  0.569±0.013 &
  0.756±0.015 &
  0.756±0.015 &
  0.766±0.056 \\
SpatialScope \cite{SpatialScope}&
  0.612±0.143 &
  0.582±0.183 &
  0.637±0.031 &
  0.683±0.114 &
  0.547±0.103 &
  0.733±0.183 &
  0.563±0.056 &
  0.769±0.022 &
  0.776±0.006 &
  0.803±0.014 \\
stDiff \cite{stDiff}&
  0.482±0.021 &
  0.527±0.013 &
  0.621±0.007 &
  0.601±0.043 &
  0.471±0.009 &
  0.544±0.021 &
  0.553±0.014 &
  0.629±0.011 &
  0.604±0.019 &
  0.736±0.099 \\
  SpaDiT \cite{li2024spadit} &
    0.657±0.035 &
  \textbf{0.621±0.099} &
    0.770 ±0.043 &
    0.725±0.106 &
    0.573±0.083 &
    0.772±0.057 &
    0.590±0.146 &
    0.808±0.043 &
    0.812±0.039 &
  0.784±0.096 \\ 
  \rowcolor{gray!20}\textbf{CausalGeD(Ours)} &
  \textbf{0.836±0.034} &
   0.612±0.043 &
  \textbf{0.825±0.052} &
  \textbf{0.853±0.043} &
  \textbf{0.918±0.066} &
  \textbf{0.921±0.050} &
  \textbf{0.912±0.096} &
  \textbf{0.877±0.020} &
  \textbf{0.879±0.009} &
  \textbf{0.895±0.127}\\
 \midrule
SSIM$\uparrow$ &
  MG &
  MH &
  MHPR &
  MVC &
  MHM &
  HBC &
  ME &
  MPMC &
  MC &
  ML \\
  \midrule
Tangram \cite{Tangram}&
  0.355±0.114 &
  0.541±0.203 &
  0.681±0.025 &
  0.653±0.115 &
  0.388±0.109 &
  0.656±0.007 &
  0.521±0.047 &
 \textbf{0.889±0.043} &
  0.789±0.004 &
  0.689±0.005 \\
scVI \cite{scVI}&
  0.487±0.155 &
  0.422±0.128 &
  0.647±0.121 &
  0.564±0.025 &
  0.374±0.115 &
  0.617±0.028 &
  0.587±0.013 &
  0.674±0.012 &
  0.736±0.006 &
  0.694±0.014 \\
SpaGE \cite{SpaGE}&
  0.503±0.003 &
  0.403±0.158 &
  0.631±0.011 &
  0.611±0.004 &
  0.401±0.006 &
  0.588±0.189 &
  0.513±0.064 &
  0.653±0.011 &
  0.667±0.055 &
  0.703±0.023 \\
stPlus \cite{stPlus}&
  0.533±0.114 &
  0.367±0.127 &
  0.657±0.176 &
  0.656±0.007 &
  0.426±0.013 &
  0.638±0.221 &
  0.479±0.023 &
  0.627±0.103 &
  0.693±0.011 &
  0.736±0.014 \\
SpaOTsc \cite{SpaOTsc}&
  0.547±0.126 &
  0.503±0.013 &
  0.701±0.026 &
  0.637±0.021 &
  0.484±0.170 &
  0.626±0.118 &
  0.601±0.188 &
  0.663±0.114 &
  0.718±0.004 &
  0.688±0.007 \\
novoSpaRc \cite{novoSpaRc}&
  0.587±0.028 &
  0.537±0.026 &
  0.713±0.123 &
  0.631±0.018 &
  0.477±0.201 &
  0.633±0.107 &
  0.622±0.023 &
  0.726±0.055 &
  0.726±0.006 &
  0.705±0.006 \\
SpatialScope \cite{SpatialScope}&
  0.612±0.016 &
  0.588±0.014 &
  0.731±0.054 &
  0.674±0.026 &
  0.512±0.122 &
  0.659±0.055 &
  0.701±0.022 &
  0.826±0.014 &
  0.753±0.014 &
  0.714±0.003 \\
stDiff \cite{stDiff}&
  0.463±0.017 &
  0.548±0.118 &
  0.673±0.013 &
  0.576±0.007 &
  0.462±0.017 &
  0.514±0.012 &
  0.563±0.017 &
  0.598±0.019 &
  0.701±0.023 &
  0.688±0.017 \\
   SpaDiT\cite{li2024spadit} &
   0.632±0.037 &
  0.574±0.125 &
  0.738±0.044 &
   0.689±0.114 &
  0.495±0.175 &
  0.717±0.111 &
  0.688±0.144 &
  0.781±0.050 &
  0.787±0.042 &
  0.751±0.107 \\ 
    \rowcolor{gray!20}\textbf{CausalGeD (Ours)} &
  \textbf{0.796±0.033} &
   \textbf{0.578±0.049} &
  \textbf{0.791±0.055} &
  \textbf{0.814±0.047} &
  \textbf{0.886±0.109} &
  \textbf{0.895±0.061} &
  \textbf{0.852±0.184} &
    0.843±0.024 &
  \textbf{0.841±0.011} &
  \textbf{0.872±0.129}\\
  \midrule
RMSE$\downarrow$ &
  MG &
  MH &
  MHPR &
  MVC &
  MHM &
  HBC &
  ME &
  MPMC &
  MC &
  ML \\ 
  \midrule
Tangram \cite{Tangram}&
  1.263±0.053 &
  1.412±0.018 &
  1.263±0.012 &
  1.587±0.041 &
  1.237±0.005 &
  1.542±0.003 &
  1.633±0.004 &
  1.324±0.048 &
  1.216±0.184 &
  1.346±0.015 \\
scVI \cite{scVI}&
  1.155±0.012 &
  1.363±0.026 &
  1.374±0.026 &
  1.327±0.106 &
  1.213±0.103 &
  1.378±0.005 &
  1.581±0.013 &
  1.207±0.034 &
  1.179±0.067 &
  1.411±0.056 \\
SpaGE \cite{SpaGE}&
  1.187±0.025 &
  1.433±0.037 &
  1.287±0.029 &
  1.354±0.047 &
  1.347±0.025 &
  1.413±0.101 &
  1.553±0.024 &
  1.137±0.011 &
  1.213±0.005 &
  1.233±0.008 \\
stPlus \cite{stPlus}&
  1.254±0.003 &
  1.367±0.045 &
  1.384±0.121 &
  1.289±0.022 &
  1.156±0.014 &
  1.331±0.077 &
  1.496±0.033 &
  1.656±0.007 &
  1.154±0.024 &
  1.303±0.014 \\
SpaOTsc \cite{SpaOTsc}&
  1.433±0.058 &
  1.213±0.058 &
  1.203±0.027 &
  1.253±0.007 &
  1.227±0.058 &
  1.203±0.114 &
  1.403±0.004 &
  1.227±0.026 &
  1.016±0.007 &
  1.263±0.005 \\
novoSpaRc \cite{novoSpaRc}&
  1.275±0.143 &
  1.526±0.213 &
  1.252±0.011 &
  1.206±0.014 &
  1.412±0.117 &
  1.198±0.007 &
  1.556±0.021 &
  1.334±0.015 &
  0.967±0.153 &
  1.523±0.007 \\
SpatialScope \cite{SpatialScope}&
  1.019±0.022 &
  1.288±0.258 &
  1.201±0.003 &
 \textbf{1.009±0.007} &
  1.217±0.005 &
  1.102±0.005 &
  1.483±0.007 &
  1.104±0.056 &
 \textbf{0.863±0.004}&
  1.343±0.014 \\
stDiff \cite{stDiff}&
  1.326±0.019 &
  1.325±0.022 &
  1.081±0.013 &
  1.219±0.066 &
  1.312±0.007 &
  1.217±0.023 &
  1.561±0.023 &
  1.326±0.016 &
  1.224±0.003 &
  1.223±0.009 \\
   SpaDiT\cite{li2024spadit} &
  \textbf{0.877±0.049} &
  \textbf{1.103±0.015} &
   1.184±0.058 &
  1.116±0.038 &
   1.125±0.060 &
  \textbf{0.992±0.045} &
    1.376±0.118 &
   1.089±0.038 &
  1.004±0.037 &
  1.121±0.047 \\ 
  \rowcolor{gray!20}\textbf{CausalGeD (Ours)} &
   1.155±0.043 &
  1.275±0.024 &
  \textbf{1.166±0.064} &
    1.115±0.040 &
  \textbf{1.053±0.037} &
    1.052±0.036 &
  \textbf{1.070±0.068} &
  \textbf{1.071±0.024} &
     1.085±0.016 &
  \textbf{1.087±0.059}\\
  \midrule
JS$\downarrow$ &
  MG &
  MH &
  MHPR &
  MVC &
  MHM &
  HBC &
  ME &
  MPMC &
  MC &
  ML \\
  \midrule
Tangram \cite{Tangram}&
  0.477±0.057 &
  0.254±0.003 &
  0.458±0.033 &
  0.343±0.007 &
  0.502±0.056 &
  0.397±0.105 &
  0.803±0.026 &
  0.403±0.056 &
  0.547±0.005 &
  0.347±0.014 \\
scVI \cite{scVI}&
  0.426±0.088 &
  0.324±0.147 &
  0.496±0.011 &
  0.403±0.001 &
  0.537±0.113 &
  0.427±0.089 &
  0.749±0.015 &
  0.423±0.115 &
  0.601±0.014 &
  0.363±0.047 \\
SpaGE \cite{SpaGE}&
  0.437±0.054 &
  0.272±0.023 &
  0.511±0.007 &
  0.387±0.114 &
  0.528±0.007 &
  0.415±0.026 &
  0.882±0.003 &
  0.374±0.004 &
  0.617±0.006 &
  0.403±0.011 \\
stPlus \cite{stPlus}&
  0.481±0.146 &
  0.288±0.057 &
  0.503±0.014 &
  0.399±0.005 &
  0.488±0.125 &
  0.439±0.005 &
  0.814±0.036 &
  0.393±0.005 &
  0.576±0.004 &
  0.423±0.016 \\
SpaOTsc \cite{SpaOTsc}&
  0.513±0.126 &
  0.334±0.058 &
  0.411±0.022 &
  0.403±0.147 &
  0.503±0.111 &
  0.411±0.015 &
  0.792±0.007 &
  0.417±0.011 &
  0.463±0.026 &
  0.311±0.007 \\
novoSpaRc \cite{novoSpaRc}&
  0.488±0.003 &
  0.401±0.017 &
  0.389±0.005 &
  0.412±0.003 &
  0.496±0.015 &
  0.429±0.085 &
  0.683±0.015 &
  0.401±0.005 &
  0.431±0.005 &
  0.401±0.006 \\
SpatialScope \cite{SpatialScope}&
  0.403±0.002 &
  0.263±0.174 &
  0.366±0.007 &
  0.389±0.008 &
  0.487±0.026 &
  0.455±0.002 &
  0.622±0.150 &
  0.389±0.107 &
  0.407±0.014 &
  0.355±0.014 \\
stDiff \cite{stDiff}&
  0.467±0.001 &
  0.412±0.015 &
  0.387±0.021 &
  0.461±0.011 &
  0.467±0.021 &
  0.456±0.011 &
  0.663±0.017 &
  0.436±0.022 &
  0.432±0.063 &
  0.396±0.007 \\
   SpaDiT\cite{li2024spadit} &
   0.346±0.012 &
   0.246±0.005 &
   0.337±0.010 &
  0.369±0.029 &
   0.463±0.116 &
   0.381±0.061 &
    0.549±0.134 &
   0.356±0.012 &
   0.371±0.013 &
  0.421±0.064 \\ 
    \rowcolor{gray!20} \textbf{CausalGeD (Ours)} &
  \textbf{0.247±0.005} &
  \textbf{0.235±0.007} &
  \textbf{0.235±0.034} &
  \textbf{0.229±0.012} &
  \textbf{0.111±0.033} &
  \textbf{0.128±0.020} &
  \textbf{0.101±0.031} &
  \textbf{0.208±0.008} &
  \textbf{0.214±0.006} &
  \textbf{0.148±0.028}\\ 
  \bottomrule
\end{tabular}%
}
\label{tab:metric_result}
\end{table*}

\begin{figure*}
    \centering
    \includegraphics[width=\textwidth]{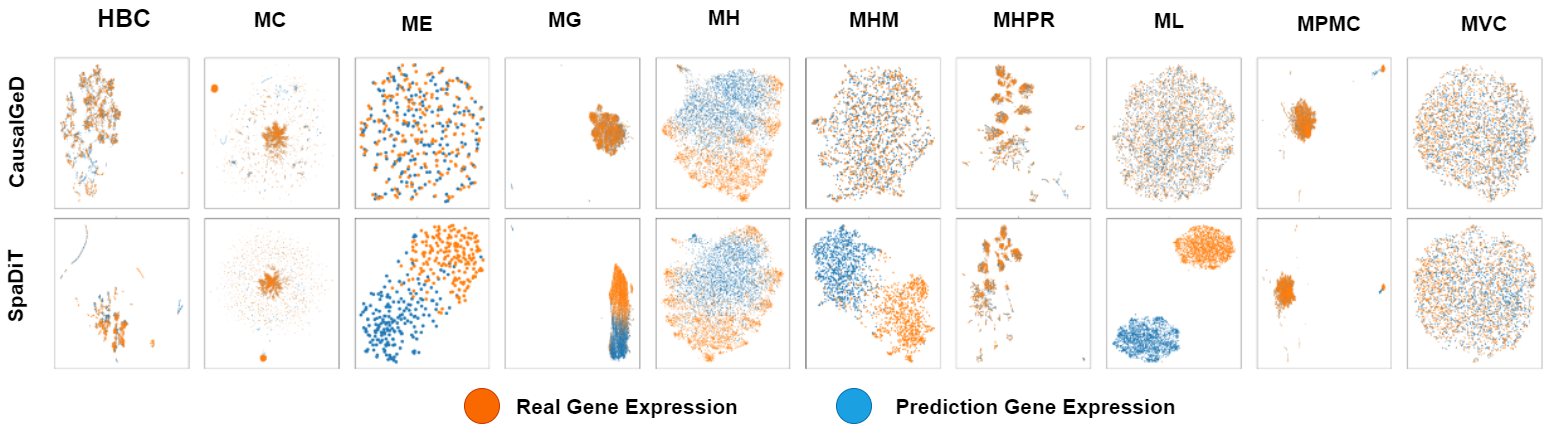}
    \caption{Low-dimensional UMAP visualizations of predicted gene expression vs. real gene expression for the proposed method and the baseline best method. CausalGeD predictions (blue) closely match real gene expression data (orange) with minimal discrepancies across ten datasets, while the baseline method shows notable discrepancies.}

    \label{fig:umap}
\end{figure*}

\subsection{Dataset and Pre-processing}
Ten benchmark datasets—scRNA sequencing and ST data—from diverse creatures' tissues were used in this study. These datasets use different sequencing platforms and technologies and come from a variety of biological organizations. Additionally, they differ in terms of missing data rates, sample sizes, and the number of geographically measured genes. In particular, 10X Chromium, Smart-seq, and Smart-seq2 are the sequencing platforms used for single-cell data in these datasets. The platforms for ST data are Slide-seqV2, 10X Visium, STARmap, MERFISH, and seqFISH. These datasets come from a variety of biological tissues, mostly tissue samples from human and mouse breast cancer.
\begin{figure*}[htbp] 
    \centering
    \includegraphics[width=\linewidth]{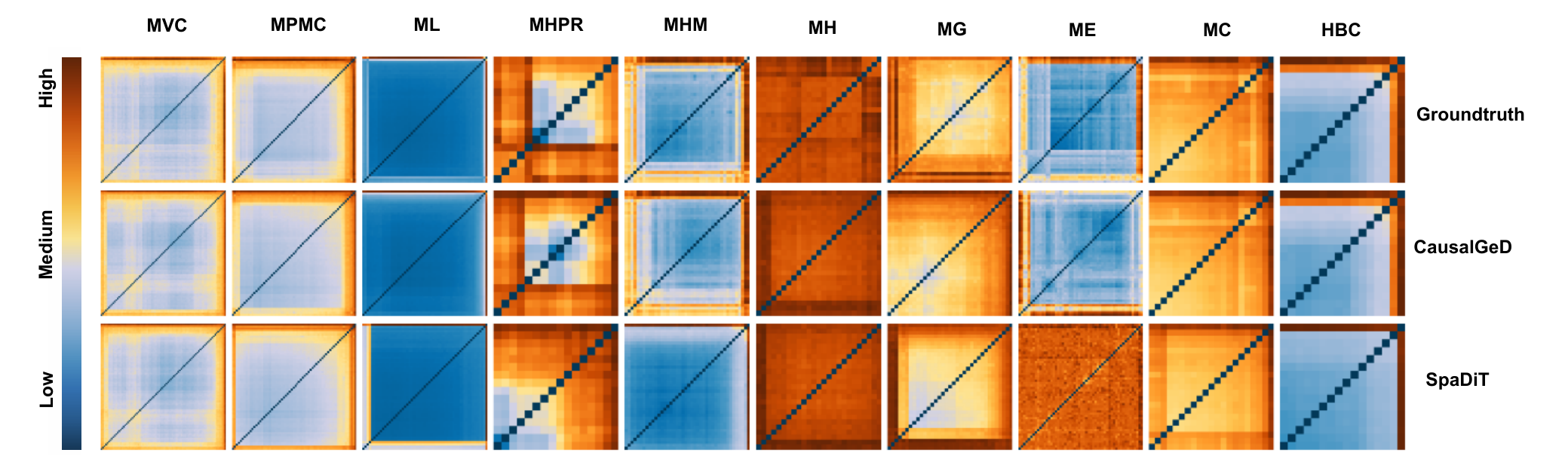} 
    \caption{ Hierarchical clustering method to visualize the similarity between the predicted genes and the true gene labels
 in the form of a heat map.}
 
    \label{fig:gene to gene}
\end{figure*}

Each dataset underwent preprocessing using the following protocol:
\paragraph{Exclusion of Low-Quality Cells} To ensure data quality, cells in the scRNA-seq datasets with fewer than 500 detected genes were removed. Similarly, for ST datasets, spots expressing fewer than one gene were excluded.
\paragraph{Expression Matrix Normalization} ST data required normalization of the expression matrices. The normalization was achieved using the following formula:
\begin{equation}
D_{ij} = \log \left( N \cdot \frac{C_{ij}}{\sum_j C_{ij}} + 1 \right),
\end{equation}
where $C_{ij}$ denotes the raw count of gene $j$ in spot $i$, and $D_{ij}$ represents the normalized read count for the same gene and spot. $N$ corresponds to the median number of transcripts detected per cell. 

\paragraph{Selection of Highly Variable Genes} For downstream analyses, genes with the highest expression variability were selected. Specifically, the top 25\% of genes demonstrating significant variation were identified, as these are likely to hold greater biological importance and better represent the underlying biological phenomena.
The processed data was divided into three distinct subsets: training, validation, and testing, maintaining a ratio of 70\%, 20\%, and 10\%, respectively. These divisions were constructed to ensure no overlap, with the test data remaining entirely independent of the training data. Results presented in this study were exclusively derived from the test set performance. \\
Dropout events in ST arise due to limitations in detecting low-expression genes, RNA content variations, and sequencing biases, leading to high data sparsity. Among the ten dataset pairs, most exhibit significant sparsity, except for the MH dataset (6.7\%), which we downsampled to create controlled high-sparsity scenarios. Model performance was evaluated using a separate test set, ensuring unbiased assessment, while hyperparameters were tuned based on validation metrics to prevent overfitting.

\subsection{Implementation Details}
The CausalGeD model adopts the CAT, which is intrinsically grounded in the Transformer architecture. Specifically, the model incorporates a dual-headed encoder to transform the input data into latent vector representations, which are subsequently subjected to a forward diffusion process by incrementally introducing noise into the latent space. The framework further integrates autoregressive modeling with the diffusion process, leading to a Causality-Aware Transformer to enhance performance. The evaluation of the model, including baseline comparisons and experimental setups across various datasets, was conducted using optimized hyperparameters. The majority of experiments were executed on NVIDIA A100 GPUs. We utilized PyTorch~\cite{paszke2019pytorch} framework to implement our work. 
We compared CausalGeD's performance with nine baseline methods, SpaDiT\cite{li2024spadit}, Tangram\cite{Tangram}, scVI\cite{scVI}, SpaGE\cite{SpaGE}, stPlus\cite{stPlus}, SpaOTsc\cite{SpaOTsc}, novoSpaRc\cite{novoSpaRc}, SpatialScope\cite{SpatialScope}, and stDiff\cite{stDiff}, using consistent data processing procedures.
\subsection{Evaluation Metrics}
We assess the performance of CausalGeD alongside the leading baseline method, SpaDiT, by employing a set of evaluation metrics to measure gene prediction accuracy across ten datasets. The Pearson Correlation Coefficient (PCC) measures the linear correlation between the ground truth and the predicted spatial expression vectors, with higher values indicating superior prediction accuracy. The Structural Similarity Index Measure (SSIM) evaluates the similarity between the ground truth and predicted spatial expression, with higher values indicating better prediction accuracy. The Root Mean Square Error (RMSE) quantifies the average difference between the z-scores of the ground truth and predicted spatial expressions, with lower values indicating better prediction accuracy. The Jensen-Shannon (JS) divergence quantifies the dissimilarity between the ground truth and predicted probability distributions of gene spatial expression, with lower values indicating closer alignment.
\subsection{Experimental Results}
Using the four metrics we evaluated both the baseline methods and CausalGeD. We reported the mean and variance of these metrics for each gene across all datasets. The outcomes are given in Table ~\ref{tab:metric_result}. The results indicate that, across the ten datasets analyzed, CausalGeD consistently delivers state-of-the-art (SOTA) performance for all four metrics. However, there are instances where CausalGeD lags behind certain established methods on specific metrics. These variations offer meaningful insights into potential areas for improving CausalGeD’s performance.

To demonstrate the effectiveness of CausalGeD in predicting gene expression while preserving both global and local structural features, we performed a comparative analysis using UMAP visualization. Figure \ref{fig:umap} presents the results across ten datasets, showing that CausalGeD predictions (blue) closely align with actual gene expression data (orange), exhibiting minimal discrepancies. In contrast, the existing best baseline SpaDiT fails to capture structural features accurately, displaying notable discrepancies. This UMAP analysis further highlights CausalGeD’s ability to maintain data integrity and effectively represent complex biological patterns.

To comprehensively showcase the precision of the CausalGeD approach in forecasting gene expression, we utilized hierarchical clustering to illustrate the resemblance between the predicted genes and the actual gene labels, and contrasted these findings with those derived from other established methods.

Initially, we computed the Euclidean distance between every pair of genes in the gene expression matrix predicted by each method, which serves as an indicator of the similarity in expression patterns between two genes: the smaller the distance, the greater the similarity. After determining the distances for all gene pairs, we applied hierarchical clustering to arrange these genes in such a way that those within the same cluster exhibit the highest similarity. This sorting allowed us to reorder the rows and columns of the distance matrix, ensuring that similar genes are positioned next to each other in the heat map.

As depicted in Figure ~\ref{fig:gene to gene}, the first row of the figure shows the true gene labels after clustering. The closer the predicted gene heat map is to these true labels, the higher the method's prediction accuracy. The figure clearly indicates that the CausalGeD method's predictions closely align with the true labels, thereby underscoring its superior accuracy in gene expression prediction.
The improved performance of CausalGeD has direct biological implications. In the HBC (human breast cancer) dataset, our model achieved a 21.8\% improvement in structural similarity, allowing better identification of spatial patterns in tumor microenvironments. This increased accuracy may help identify previously unclear cell-cell communication patterns and regulatory networks within the tumor ecosystem. Similarly, in the mouse embryo (ME) dataset, the higher prediction accuracy (PCC improvement of 34.9\%) may enable better understanding of developmental gene regulation patterns, which is crucial to studying embryonic development and tissue formation. These improvements in prediction accuracy help translate into more reliable insights into biological processes such as cell differentiation, tissue organization, and disease progression.
\section{Ablation Studies}
To evaluate the efficacy of our proposed approach, we performed comprehensive ablation studies on various components of CausalGeD. This included examining the flexibility of the VAE decoder during training, variation in encoder training, different transformer blocks, diffusion time steps, and analyzing the impact of different AR steps with decayed sampling strategies.
\subsection{Decoder Training Options}
\begin{figure}[h]
    \centering
    \includegraphics[width=\linewidth, trim={0.09cm 0.09cm 0.09cm 0.09cm}, clip]{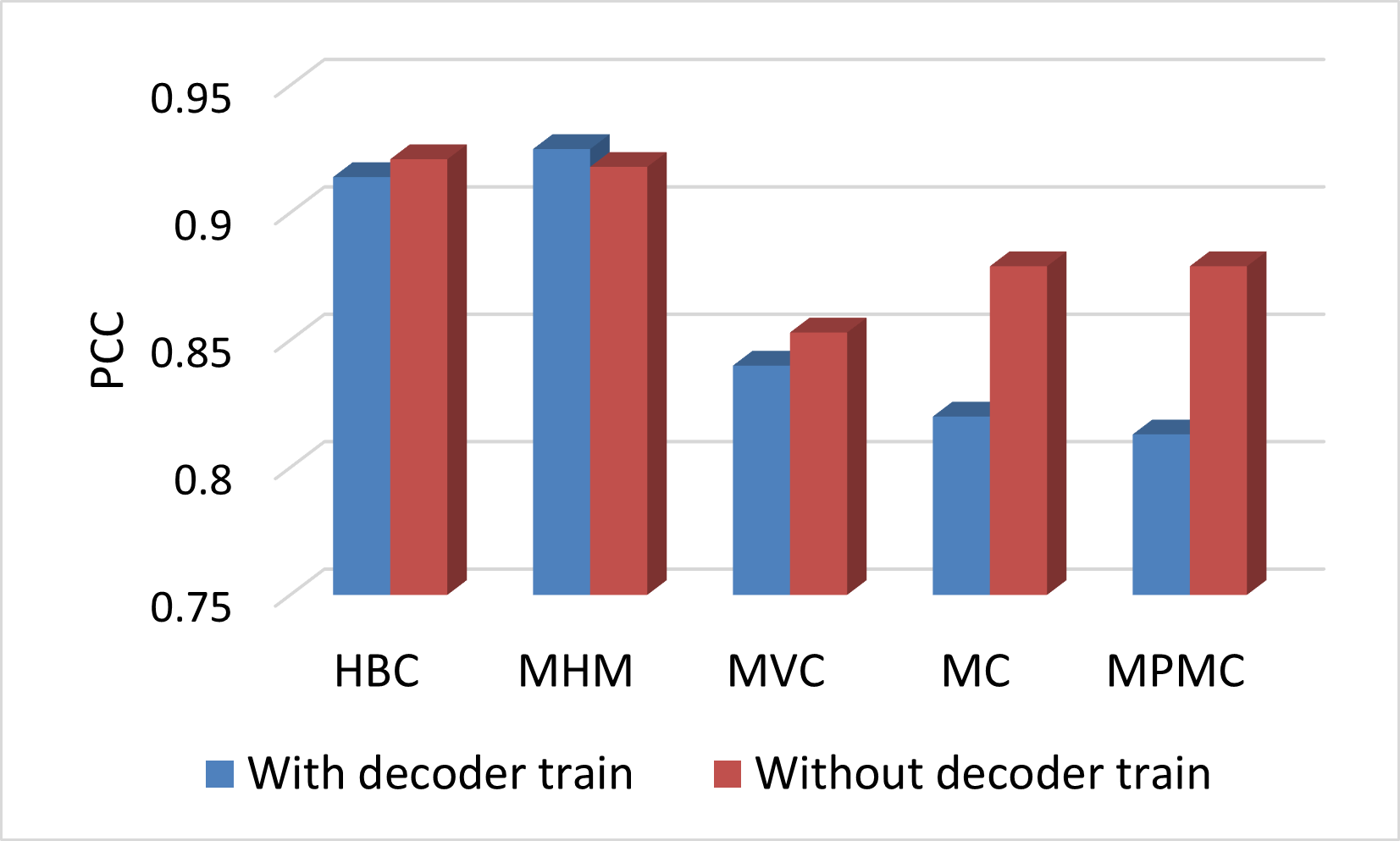}
    \caption{Ablation study of decoder training module}
    \label{fig:decoder_train}
\end{figure}

In our ablation study, we investigated the impact of training the VAE decoder in conjunction with the CausalGeD model on various datasets, as illustrated in Figure \ref{fig:decoder_train}. In the HBC dataset, adding decoder training slightly reduced PCC from 0.921 to 0.914 (a 0.7\% decrease). Decoder training shows varying effects across datasets: it improves PCC from 0.918 to 0.925 in the MHM dataset, but decreases PCC from 0.853 to 0.840 in MVC. For MC and MPMC datasets, training without the decoder yields better results, improving PCC from 0.820 to 0.879 and from 0.813 to 0.879, respectively. These findings suggest that skipping decoder training generally improves performance, though the magnitude of improvement varies across datasets, highlighting the importance of dataset-specific training strategies.
\subsection{Encoder Variation}
\begin{figure}[htbp] 
    \centering
    \includegraphics[width=\linewidth, trim={0.09cm 0.09cm 0.09cm 0.09cm}, clip]{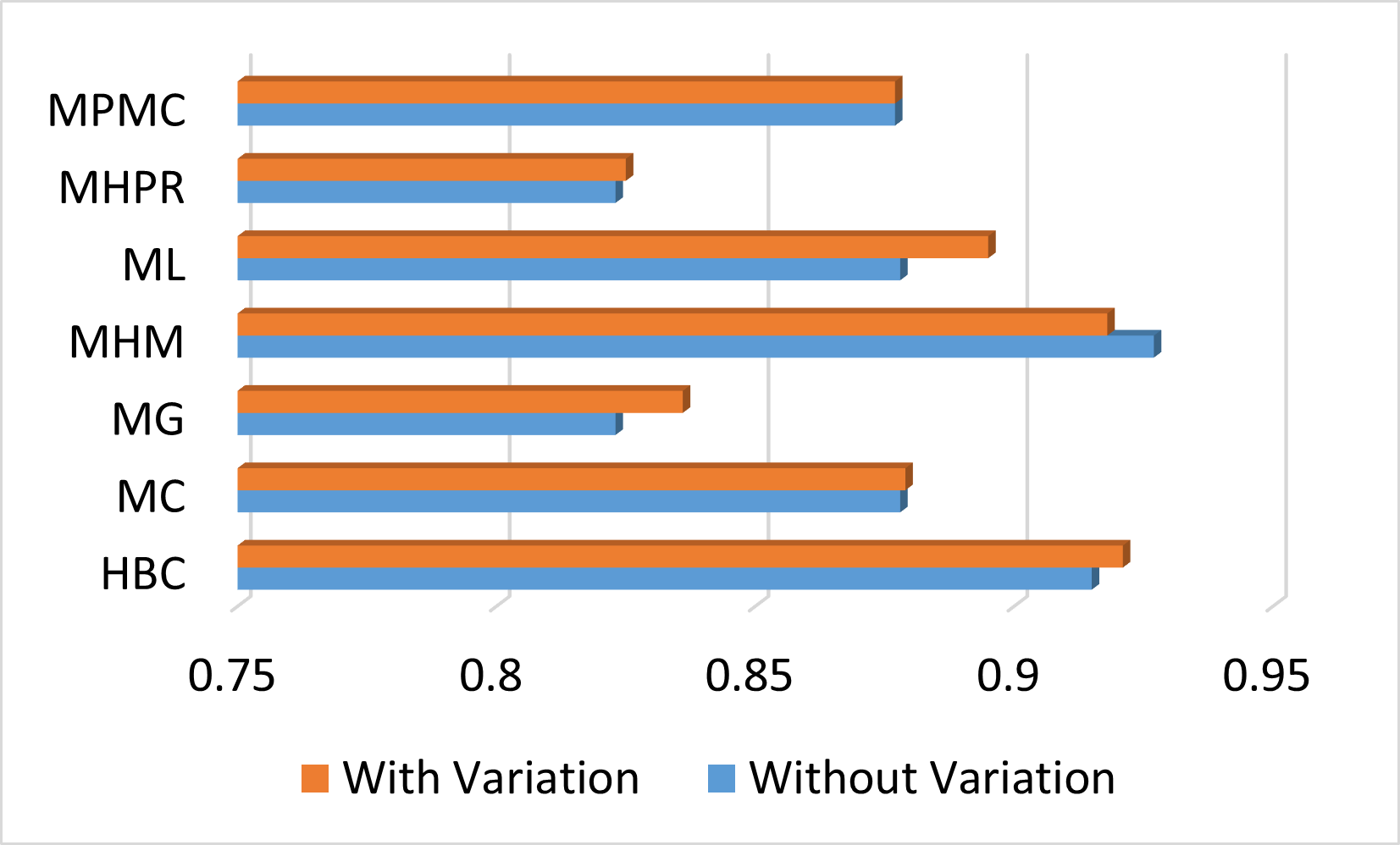} 
    \caption{ Comparison of PCC values across different datasets with and without encoder variation.}

    \label{fig:encoder_variation}
\end{figure}

We also investigated the encoder with and without variation training in our CausalGeD model training to assess the variation autoencoder. The results, in Figure \ref{fig:encoder_variation} represented by orange bars for training with encoder variation and blue bars for training without it, indicate that encoder variation generally leads to slight improvements in PCC values across most datasets. Notably, the MG dataset shows the most significant increase, with PCC improving from 0.823 to 0.836, while the ML dataset also benefits from encoder variation, increasing from 0.878 to 0.895. However, a slight performance drop is observed in the MHM dataset, where the PCC decreases from 0.927 to 0.918. Overall, the improvements observed in several datasets suggest that encoder variation enhances feature extraction and generalization, although minor declines in specific datasets indicate the need for further fine-tuning to optimize performance.
\subsection{AR Step Decay}  
We investigated how gradually reducing the influence of AR steps affects model performance. The AR step decay controls how much weight we give to each successive step in the sequence - similar to how earlier regulatory genes have stronger influence than later ones in biological pathways.

\begin{figure}[htbp] 
    \centering
    \includegraphics[width=0.9\linewidth]{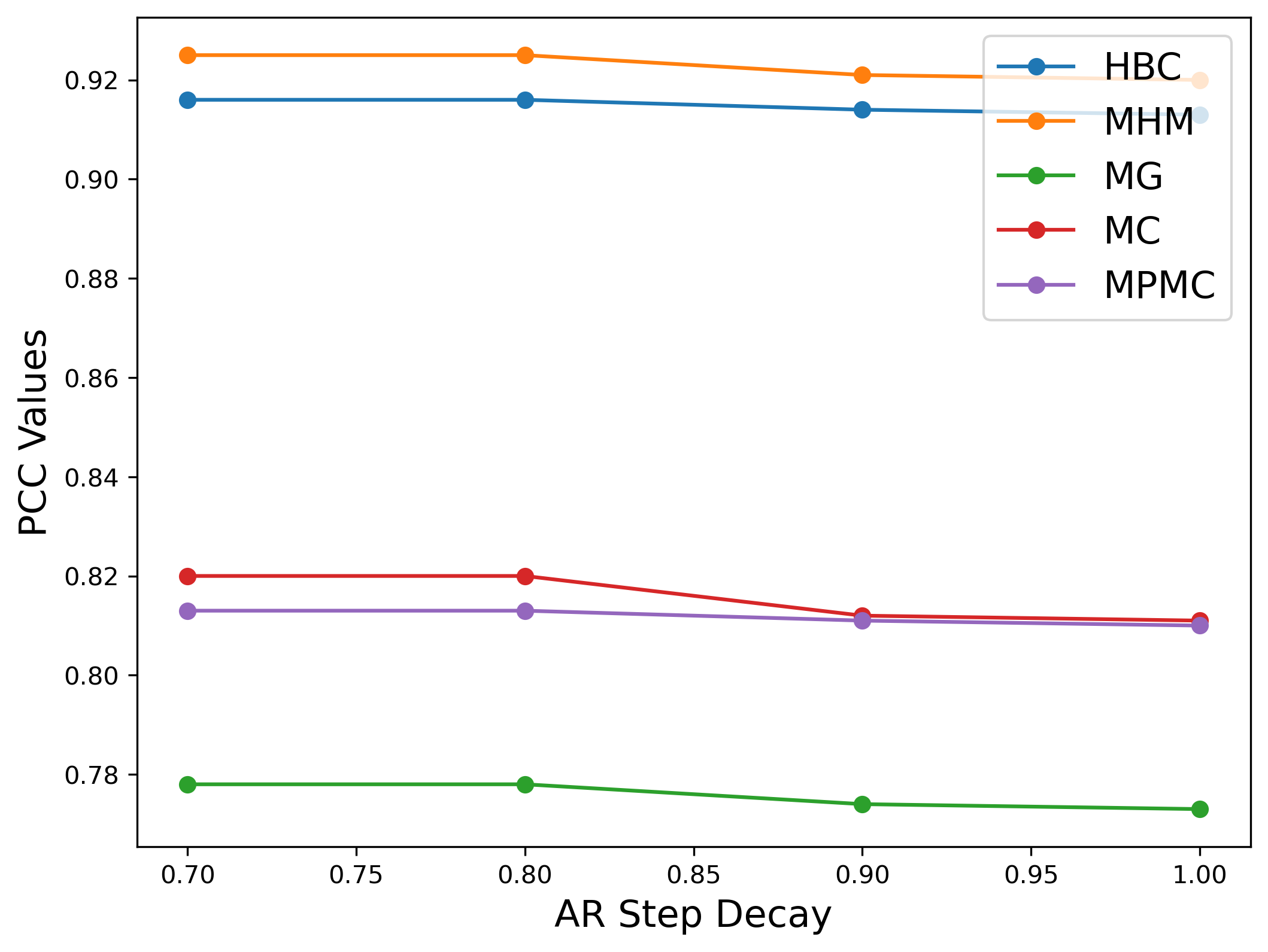} 
    \caption{The effect of AR step decay}
  
    \label{fig:arstep}
\end{figure}

As shown in Figure \ref{fig:arstep}, increasing the decay rate from 0.7 toward 1.0 (making steps more equal in influence) leads to consistent decreases in prediction accuracy (e.g., PCC values) across all datasets. This suggests that our biologically inspired decaying influence pattern, where earlier steps have more impact than later ones, better captures the hierarchical nature of gene regulatory relationships. The decay pattern is particularly effective for datasets with complex regulatory networks, as evidenced by the more stable performance in HBC, MC, MG and MHM datasets where known regulatory hierarchies exist. Our analysis suggests that an optimal decay ratio of 0.8 balances generalization and performance across different datasets.

\subsection{Transformer Blocks}
\begin{figure}[htbp] 
    \centering    \includegraphics[width=0.9\linewidth, trim={0.09cm 0.09cm 0.09cm 0.09cm}, clip]{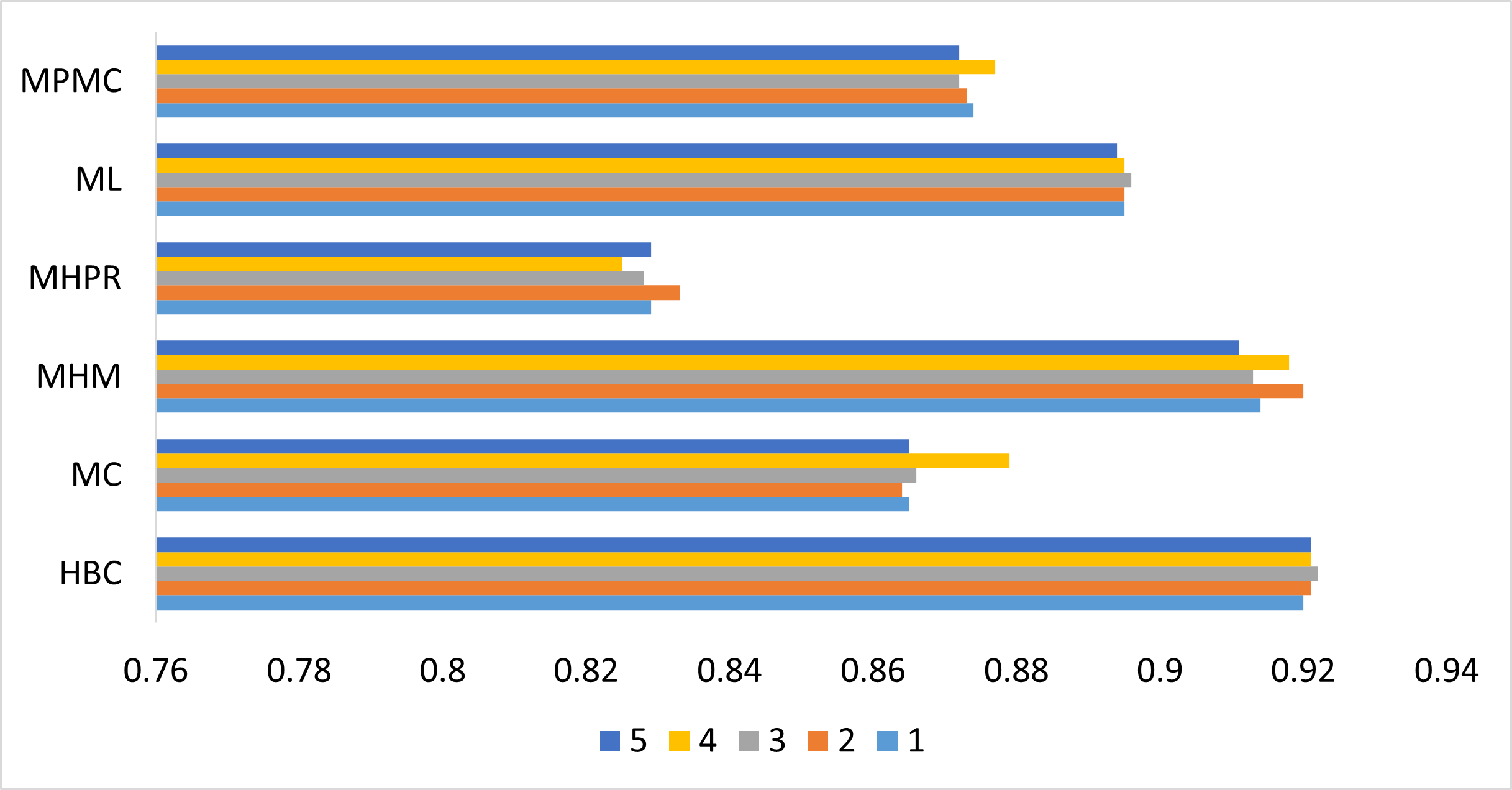} 
    \caption{Effect of varying transformer blocks on PCC values across datasets}    
    \label{fig:depth}
\end{figure}

In our CausalGeD model we investigated the impact of transformer block depth on performance-efficiency trade-offs, dataset-specific variability, and generalization robustness across diverse datasets. Figure \ref{fig:depth} illustrates the performance of the CausalGeD model across different datasets (y-axis) using varying numbers of transformer blocks (x-axis), indicated by 1, 2, 3, 4, and 5. The performance metric, PCC (Pearson Correlation Coefficient), remains consistently high across all datasets, with minor variations as the number of transformer blocks increases. Notably, using 3 or more transformer blocks appears to stabilize and slightly enhance performance, as seen by the convergence of PCC values close to the upper range (\(\sim 0.92 - 0.94\)). This suggests that for CausalGeD, a model depth of at least 3 transformer blocks provides a good balance between complexity and performance, ensuring robust and reliable predictions across diverse datasets.
\subsection{Diffusion Time Steps}
To investigate the robustness of the CausalGeD model we explore the impact of using different T values for different AR steps during training. The results in Table \ref{tab:timestep} demonstrate that varying the diffusion time step sampling strategy does not significantly impact model performance, as indicated by the consistent PCC values across different datasets. The default setting (T=2000) yields results comparable to other strategies, such as $1/20 T$, $1/4 T$, $1/2 T$, and $1/3 T$, with minimal variations observed. This consistency highlights the robustness of the CausalGeD model to different diffusion time step configurations, suggesting that the model performs reliably regardless of whether a shared or random time step strategy is used.
\begin{table}[h]
    \centering
    \caption{Robustness on diffusion time steps sampling strategy.}
 \resizebox{\linewidth}{!}{%
    \begin{tabular}{lccccccc}
        \toprule
        Time Steps & HBC & MC & MG & MHM & MHPR & ML & MPMC\\
        \midrule
         T (Default) & 0.921  & 0.879  & 0.836  & 0.918  & 0.825  & 0.895  & 0.877 \\ 
         \midrule
        1/20 T  & 0.921  & 0.864  & 0.835  & 0.913  & 0.831  & 0.895  & 0.872 \\
        \midrule
        1/4 T  & 0.921  & 0.864  & 0.835  & 0.920  & 0.827 & 0.894  & 0.872 \\ 
        \midrule
        1/2 T  & 0.921 & 0.864  & 0.835  & 0.920 & 0.827 & 0.894  & 0.872 \\ 
        \midrule
        1/3 T  & 0.920  & 0.865  & 0.836  & 0.913  & 0.831  & 0.895  & 0.872 \\ 
        \bottomrule
    \end{tabular}%
    } 
    \label{tab:timestep}
\end{table}

\section{Conclusion}
We introduce CausalGeD, a novel framework that integrates diffusion models with causal relationships for spatial transcriptomics and single-cell RNA sequencing data integration. Our approach outperforms the latest baselines by 5- 32\% in key metrics in ten tissue datasets, demonstrating that modeling gene regulatory relationships significantly improves spatial gene expression prediction. Beyond technical performance, CausalGeD provides new insights into gene regulatory mechanisms in spatial contexts, advancing both computational capabilities and biological understanding.

A potential limitation of our current model is that the ST genes have to be a subset of the genes in the scRNA-seq. Otherwise, our approach might not be directly applied. Extension to more general scenarios is our future research line.  
\section*{Acknowledgment}
This research is supported in part by the NSF under Grant IIS 2327113 and ITE 2433190 and the NIH under Grants R21AG070909 and P30AG072946. We would like to thank NSF for support for the AI research resource with NAIRR240219. We thank the University of Kentucky Center for Computational Sciences and Information Technology Services Research Computing for their support and use of the Lipscomb Compute Cluster and associated research computing resources. We also acknowledge and thank those who created, cleaned, and curated the datasets used in this study. 

\bibliographystyle{IEEEtran}
\bibliography{sample-base}
\section{Appendix}
\subsection{Causal Mask formation}
\label{sec:causal_mask}
\begin{figure*}[ht]
    \centering
    \subfloat[Conditional Tokens]{%
        \includegraphics[width=100pt, height=100pt]{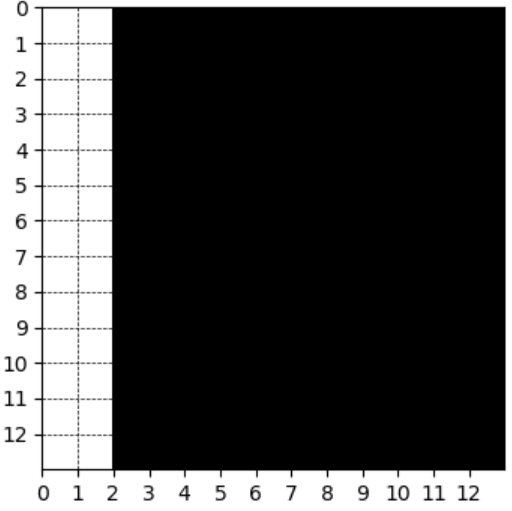}
    }
    \hfill
    \subfloat[Visible to Visible]{%
        \includegraphics[width=100pt, height=100pt]{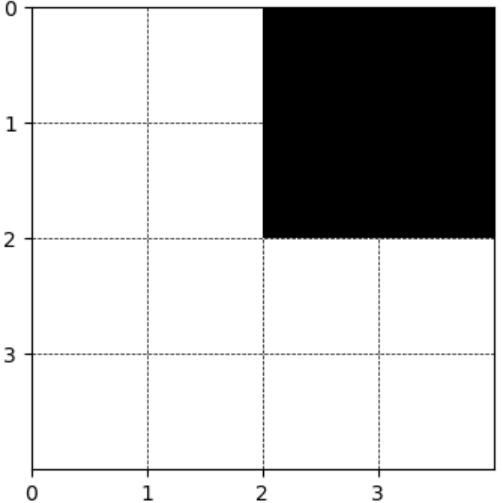}
    }
    \hfill
    \subfloat[Sample to Visible]{%
        \includegraphics[width=88pt, height=100pt]{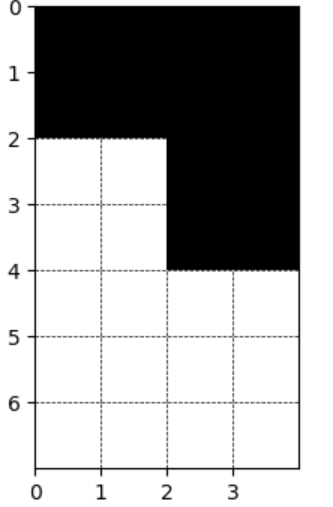}
    }
    \hfill
    \subfloat[Sample to Sample]{%
        \includegraphics[width=100pt, height=100pt]{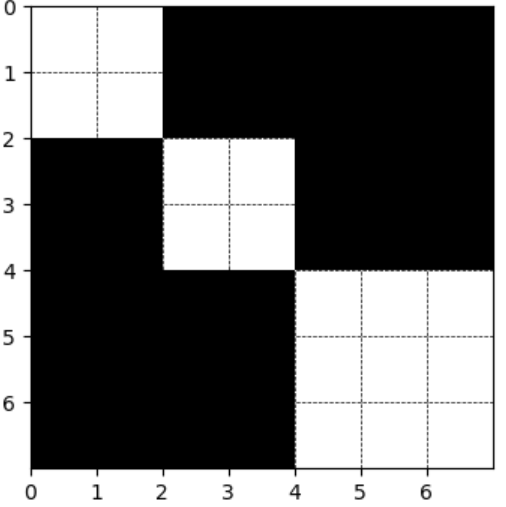}
    }
    \hfill
    \subfloat[Causal Attention Mask]{%
        \includegraphics[width=100pt, height=100pt]{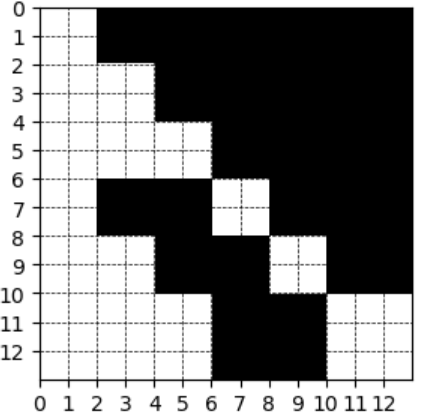}
    }

    \caption{Illustration of the different parts in the process of causal attention mask formation. White indicates allowed regions, while black indicates masked regions.}
    \label{fig:attn_mask}
\end{figure*}

The beginning of the Causal Attention mask starts with algorithm-\ref{alg:split_integer_exp_decay}. From this algorithm the split size $sz$ and the cumulative sum $cs$ are returned to algorithm \ref{alg:attn_mask}. For example, the demonstration of this formation of causal attention masks illustrated in Figure ~\ref{fig:attn_mask}. If the sample length
\( s = 7 \) and a conditional length of \( c = 2 \), with split sizes
\( sz = [2,2,3] \) and cumulative sums \( cs = [0,2,4,7] \). Key lengths are calculated: 
visible length \( v = 4 \), context length \( ctx = 6 \), and total sequence length 
\( seq = 13 \). A \( 13 \times 13 \) matrix is initialized with ones and with the first two columns set to zero to unmask conditional tokens. The mask is then partitioned into sub-matrices—visible-to-visible (\( vTv \)), sample-to-visible (\( sTv \)), and sample-to-sample (\( sTs \)).
\begin{algorithm}[H]
\caption{Generate AR Steps}
\label{alg:split_integer_exp_decay}
\textbf{Input:} $S$ (sample length to split), $\alpha$ (exponential decay factor)
\vspace{-0.4cm}
\begin{algorithmic}[1]
\IF{$\alpha = 1.0$}
    \STATE $N \gets \text{random integer in } [1, S]$
\ELSE
    \STATE $b \gets \frac{1 - \alpha}{1 - \alpha^S}$ \hfill {\color{gray}\texttt{\% Normalization factor}}
    \STATE $p \gets [b \cdot \alpha^i \text{ for } i \in [0, S-1]]$
    \STATE $N \gets \text{random choice from } [1, S] \text{ with probabilities } p$
\ENDIF

\STATE \textbf{Generate cumulative sum:}
\STATE $cs \gets [0] + \text{sort(random sample from } [1, S) \text{ of size } N-1) + [S]$
\STATE $sz \gets [cs[i+1] - cs[i] \text{ for } i \in [0, N-1]]$

\STATE \textbf{return} $sz, cs$
\end{algorithmic}
\end{algorithm}
\vspace{-0.5cm}
\begin{algorithm}[H]
\caption{Generate Causal Attention Mask}
\label{alg:attn_mask}
\textbf{Input:} $s$ (sample length), $c$ (conditional length), $sz$ (split sizes), $cs$ (cumulative sum of split sizes)
\begin{algorithmic}[1]
\STATE $v \gets s - sz[-1]$ \hfill {\color{gray}\texttt{\% Visible length}}
\STATE $ctx \gets c + v$ \hfill {\color{gray}\texttt{\% Context length}}
\STATE $seq \gets ctx + s$ \hfill {\color{gray}\texttt{\% Total sequence length}}

\STATE Initialize $\mathbf{M} \in \mathbb{R}^{seq \times seq}$ with ones
\STATE $\mathbf{M}[:,:c] \gets 0$ \hfill {\color{gray}\texttt{\% Condition tokens}}

\STATE \textbf{Build partitions:}
\STATE $vTv \gets \mathbf{1}^{v \times v}$ \hfill {\color{gray}\texttt{\% Visible to visible}}
\STATE $sTv \gets \mathbf{1}^{s \times v}$ \hfill {\color{gray}\texttt{\% Sample to visible}}
\STATE $sTs \gets \mathbf{1}^{s \times s}$ \hfill {\color{gray}\texttt{\% Sample to sample}}

\FOR{$i = 0$ \TO $|sz| - 2$}
    \STATE $vTv[cs[i]:cs[i+1], 0:cs[i+1]] \gets 0$
    \STATE $sTv[cs[i+1]:cs[i+2], 0:cs[i+1]] \gets 0$
\ENDFOR

\FOR{$i = 0$ \TO $|sz| - 1$}
    \STATE $sTs[cs[i]:cs[i+1], cs[i]:cs[i+1]] \gets 0$
\ENDFOR

\STATE \textbf{Build attention mask:}
\STATE $\mathbf{M}[c:ctx, c:ctx] \gets vTv$
\STATE $\mathbf{M}[ctx:, c:ctx] \gets sTv$
\STATE $\mathbf{M}[ctx:, ctx:] \gets sTs$

\STATE \textbf{return} $\mathbf{M}$
\end{algorithmic}
\end{algorithm}

\subsection{Dataset details}

\begin{table*}[t]
\centering
\caption{Overview of the Datasets used in the research and key statistics.}
\small
\renewcommand{\arraystretch}{1.1}
\begin{adjustbox}{max width=\linewidth}
\begin{tabular}{llccccccccccccr}
\toprule
\multirow{2}{*}{\textbf{Datasets}} & \multirow{2}{*}{\textbf{Tissue}} & \multirow{2}{*}{\textbf{GEO ID}} & \multirow{2}{*}{\textbf{SC  Platform}} & \multirow{2}{*}{\textbf{ST  Platform}} & \multicolumn{2}{c}{\textbf{Cells/Spots}} & \multicolumn{2}{c}{\textbf{Genes}} & \multicolumn{2}{c}{\textbf{Prepro. Cells/Spots}} & \multicolumn{2}{c}{\textbf{Prepro. Genes}} & \multicolumn{2}{c}{\textbf{Dropout Rate}} \\
\cline{6-15}
& & & & & SC & ST & SC & ST & SC & ST & SC & ST & SC & ST \\
\midrule
MH\cite{data-MH} & mouse hippocampus & GSE158450 & 10X Chromium & seqFISH & 8596 & 3585 & 16384 & 249 & 8584 & 3585 & 1260 & 249 & 80.3\% & 6.3\% \\
MHPR\cite{data-MHPR} & mouse HPR & GSE113576 & 10X Chromium & MERFISH & 31299 & 4975 & 18646 & 154 & 31299 & 4975 & 1939 & 153 & 73.7\% & 62.2\% \\
ML\cite{data-MG} & mouse liver & GSE109774 & Smart-seq2 & seqFISH & 981 & 2177 & 17533 & 19532 & 887 & 2177 & 2279 & 569 & 73.2\% & 75.4\% \\
MG\cite{data-MG} & mouse gastrulation & GSE15677 & 10X Chromium & seqFISH & 4651 & 8425 & 19103 & 351 & 4651 & 8425 & 1945 & 345 & 58.6\% & 74.1\% \\
MVC\cite{data-MVC} & mouse visual cortex & - & Smart-seq & STARmap & 14249 & 1549 & 34041 & 1020 & 14249 & 1549 & 3774 & 844 & 58.2\% & 76.2\% \\
\midrule
MHM\cite{data-MHM} & mouse hindlimb muscle & GSE161318 & 10X Chromium & 10X Visium & 4816 & 995 & 15460 & 33217 & 4809 & 995 & 1667 & 416 & 80.3\% & 68.9\% \\
HBC\cite{data-HBC} & human breast cancer & CID3586 & 10X Chromium & 10X Visium & 6178 & 4784 & 21164 & 28402 & 6143 & 4784 & 625 & 125 & 76.6\% & 70.6\% \\
ME\cite{data-ME} & mouse embryo & GSE160137 & 10X Chromium & 10X Visium & 3415 & 198 & 19374 & 53574 & 3415 & 198 & 2163 & 540 & 61.1\% & 62.3\% \\
MPMC\cite{data-MPMC} & mouse PMC & - & 10X Chromium & 10X Visium & 3499 & 9852 & 24340 & 24518 & 3499 & 9852 & 2544 & 636 & 70.6\% & 81.7\% \\
MC\cite{data-MC} & mouse cerebellum & SCP948 & 10X Chromium & Slide-seqV2 & 26252 & 41674 & 24409 & 23264 & 26252 & 41674 & 822 & 205 & 79.5\% & 83.9\% \\
\bottomrule
\end{tabular}
\end{adjustbox}
\label{tab:data}
\end{table*}
Table ~\ref{tab:data} illustrates the paired scRNA-seq and spatial transcriptomic datasets with corresponding tissue types, sequencing platforms, and preprocessing information. \textbf{Prepro. Cells/Spots} and \textbf{Prepro. Genes} show the number of cells/spots and genes remaining after preprocessing. The datasets also highlight differences between imaging-based ST methods (seqFISH, MERFISH, STARmap, Slide-seqV2), which generally have higher dropout rates, and sequencing-based ST methods (10X Visium), which provide broader transcriptome coverage. This integrative dataset enables spatial gene expression modeling by leveraging the spatial resolution of ST data and the higher gene detection capability of scRNA-seq, making it a valuable resource for studying tissue heterogeneity and cell-type composition across both human and mouse samples. \textbf{Dropout Rate} represents the proportion of missing values (zeros) in the data. HPR: hypothalamic preoptic region; PMC: primary motor cortex.
\end{document}